\documentclass[10pt,conference]{IEEEtran}
\usepackage[utf8]{inputenc}
\usepackage[T1]{fontenc}
\usepackage{amsmath,amssymb,amsfonts}
\usepackage{booktabs,multirow}
\usepackage{graphicx}
\usepackage{cite}
\usepackage{microtype}
\usepackage{enumitem}
\usepackage{url}
\usepackage{xcolor}
\usepackage{tikz}
\usetikzlibrary{arrows.meta,positioning,fit,calc}
\usepackage{balance}
\usepackage{hyperref}
\hypersetup{hidelinks}
\setlist[itemize]{leftmargin=*,nosep}
\setlist[enumerate]{leftmargin=*,nosep}
\definecolor{vision}{HTML}{3B82F6}
\definecolor{wireless}{HTML}{F59E0B}
\definecolor{joint}{HTML}{10B981}
\definecolor{adapt}{HTML}{8B5CF6}
\newcommand{\R}{\mathbb{R}}

\newcommand{\cL}{\mathcal{L}}
\newcommand{\cD}{\mathcal{D}}
\newcommand{\cW}{\mathcal{W}}

\title{CM-MAE: A Physics-Guided Cross-Modal Self-Supervised Learning Framework for Vision-Wireless Applications}

\author{\IEEEauthorblockN{Yubo Zhang, Yiyao Liu}}

\begin{document}
\maketitle

\begin{abstract}
Synchronized camera and wireless measurements observe the same scene through different physical channels. The central difficulty is that a representation learned in one deployment can fail when viewpoint, traffic, illumination, and propagation geometry change. This paper presents CM-MAE, a self-supervised vision--wireless pretraining framework for cross-scenario representation transfer. The evaluated real-data model uses only RGB frames and the measured 64-beam received-power vector available in DeepSense 6G; it does not use ray-traced paths, calibrated depth, or beam-index labels during pretraining. Its central pretraining term is a \emph{soft contrastive alignment loss}. Instead of making the synchronized image--wireless pair the only positive pair, this loss builds a target distribution from similarities between measured beam-power profiles, so nonidentical samples with similar directional responses are not forced apart as false negatives. A masked joint decoder provides the complementary local objective by reconstructing hidden visual patches and wireless angular clusters under modality dropout. After pretraining, a differential-rate fine-tuning rule lets a new fusion head adapt quickly while the encoders move slowly. Under a sequence-disjoint DeepSense 6G protocol, adding the soft alignment loss improves a matched linear-probe transfer average from 24.88\% to 29.49\%. Mild fusion fine-tuning reaches 77.38\% Top-1 accuracy on unseen Scenarios 6--8, and optional transductive normalization adaptation reaches 78.69\%. Since the fusion setting uses the contemporaneous 64-beam power vector at inference, these results should be read as representation-transfer diagnostics, not as proactive beam-prediction or reduced-sweeping claims.
\end{abstract}


\section{Introduction}
\label{sec:intro}

Vision-assisted wireless learning is attractive because the camera observes geometry and occlusion while the wireless receiver observes the directional response induced by the same scene. In a vehicle-to-infrastructure deployment, however, the relation between image pixels and a selected beam is not fixed. It changes with camera placement, street layout, traffic composition, reflectors, blockers, and array response. A model that fits one measured scenario can therefore lose useful structure when moved to another.

This paper focuses on the part of the problem that can be studied with real synchronized measurements: learning transferable vision--wireless representations without using discrete beam-index labels during pretraining. Each training sample contains an RGB image and a 64-dimensional received-power vector obtained from a beam sweep. The downstream task is to classify the strongest measured beam index, but that index is withheld during self-supervised pretraining.

Two design choices are needed. First, masked reconstruction should not be purely unimodal. If the image branch reconstructs only image patches and the wireless branch reconstructs only wireless features, the model can learn useful local statistics while leaving the two global representations weakly related. Second, cross-modal alignment should not be a hard instance-matching loss. CLIP-style alignment treats every nonpaired sample in the batch as a negative~\cite{clip}, even when two wireless power profiles have nearly the same dominant angular response. For beam-sweep data, this creates false negatives: samples that are different frames but physically similar from the array's perspective.

CM-MAE addresses these two issues directly. Its global objective is a \emph{soft contrastive alignment loss}: the target for a visual query is not a one-hot sample identity, but a probability distribution over wireless samples computed from measured beam-profile similarity. Its local objective is masked conditional reconstruction through a joint decoder. The evaluated model implements these ideas with camera-ray-indexed image patches and sparse one-dimensional angular clusters extracted from the 64-beam power vector. A geometry-privileged extension would require calibrated wideband CFR and ray-traced correspondence labels; because those signals are not used in the reported experiments, the extension is discussed only as a limitation and future direction.

The main contributions are as follows.
\begin{itemize}
    \item We formulate real-data CM-MAE pretraining using only synchronized RGB frames and measured beam-power vectors, with a clear separation between pretraining inputs and downstream beam-index labels.
    \item We introduce a beam-profile-guided soft contrastive alignment loss that matches learned cross-modal retrieval distributions to measured wireless-response neighborhoods.
    \item We combine this global soft alignment with masked conditional reconstruction in a dual-stream Transformer architecture whose masking, decoder, and modality-dropout rules are explicitly gated by the available modalities.
    \item Matched ablations show a 4.61-point average transfer gain from soft alignment. Differential-rate adaptation raises the unseen-scenario average from 51.36\% to 77.38\%; optional transductive adaptation reaches 78.69\%.
\end{itemize}

\section{Background and Related Work}
\label{sec:related}

\textbf{Contrastive and masked self-supervision.}
Contrastive learning builds representations by making related views close and unrelated views far apart. In cross-modal learning, CLIP-style training uses the paired sample as the positive and all other samples in the batch as negatives~\cite{clip}. This objective is simple and effective when the correct relation is sample identity, but it can be too rigid for directional wireless measurements, where multiple frames can have similar beam responses. Masked autoencoding instead hides input tokens and trains a decoder to reconstruct them~\cite{mae}. MultiMAE and CAV-MAE show that masked reconstruction can be extended to multiple modalities and combined with alignment losses~\cite{multimae,cavmae}. CM-MAE follows this general family but changes the alignment target from hard identity to a wireless-response neighborhood.

\textbf{Wireless and vision--wireless representation learning.}
DeepSense 6G provides real measured sensory and wireless data for multimodal communication tasks~\cite{deepsense}. Vision-position beam prediction demonstrates the usefulness of side information under supervised training~\cite{visionposition}. Recent self-supervised or foundation-style wireless models include LWM~\cite{lwm}, ContraWiMAE~\cite{contrawimae}, WMFM~\cite{wmfm}, WiFo-M$^2$~\cite{wifom2}, and WiFo-MiSAC~\cite{wifomisac}. These works motivate masked and multimodal learning for wireless systems. CM-MAE is narrower in input scope but sharper in target construction: for real DeepSense data, it aligns camera and wireless features using the measured 64-beam power-profile geometry available in each synchronized sample.

\textbf{Adaptation across scenarios.}
Cross-scenario evaluation is essential because source validation accuracy can overstate transfer performance. Test-time adaptation methods such as Tent update normalization parameters on unlabeled target data by minimizing prediction entropy~\cite{tent}. Domain-adversarial training seeks scenario-invariant features through gradient reversal~\cite{dann}. CM-MAE differs from these adaptation-only approaches in two ways: it first builds a cross-modal representation through beam-profile-guided pretraining, and then uses a conservative supervised adaptation rule in which new fusion layers learn quickly while pretrained encoders move slowly. In our experiments, Tent-style normalization adaptation is kept as an optional transductive stage, while the adversarial objective was numerically unstable and is not part of the final method.

\section{Problem Setup}
\label{sec:setup}

\subsection{Synchronized Image and Beam-Power Measurements}
Consider a base station equipped with a beam codebook $\cW=\{\mathbf w_b\}_{b=1}^{N_b}$ and a co-located RGB camera. A synchronized sample is denoted
\begin{align}
&\left(I_i,\boldsymbol\rho_i\right),\qquad
I_i\in\R^{3\times H\times W}, \nonumber\\
&\boldsymbol\rho_i=[\rho_{i,1},\ldots,\rho_{i,N_b}]^\top\in\R_+^{N_b}.
\label{eq:sample}
\end{align}
Here $I_i$ is the RGB frame and $\rho_{i,b}$ is the measured received power when beam $b$ is used in the sweep. DeepSense provides $N_b=64$. If $\mathbf h_{i,k}$ is the effective channel on subcarrier $k$, the sweep measurement can be written as
\begin{equation}
\rho_{i,b}=\frac{1}{K}\sum_{k=1}^{K}
\left|\mathbf w_b^{H}\mathbf h_{i,k}\right|^2+\xi_{i,b},
\quad b=1,\ldots,N_b,
\label{eq:power_model}
\end{equation}
where $\xi_{i,b}$ includes noise and measurement perturbations. The downstream class label is the strongest measured beam index
\begin{equation}
b_i^\star=\arg\max_{b\in\{1,\ldots,N_b\}}\rho_{i,b}.
\label{eq:selected_beam}
\end{equation}
Equation~\eqref{eq:selected_beam} is used for downstream supervised training and evaluation only. CM-MAE pretraining does not consume $b_i^\star$ as a label. It does consume the full measured power vector $\boldsymbol\rho_i$, which is task-relevant because it contains the maximizer in~\eqref{eq:selected_beam}.

The image is divided into $P=(H/S)(W/S)$ nonoverlapping patches of size $S\times S$. Let $\widetilde{\mathbf q}_p$ be the homogeneous image coordinate of patch center $p$ and let $K_{\rm cam}$ be the camera intrinsic matrix. The normalized camera ray used by the tokenizer is
\begin{equation}
\mathbf d_p=\frac{K_{\rm cam}^{-1}\widetilde{\mathbf q}_p}
{\|K_{\rm cam}^{-1}\widetilde{\mathbf q}_p\|_2}.
\label{eq:ray}
\end{equation}
In the evaluated DeepSense implementation, $K_{\rm cam}$ is an approximate pinhole camera with a $90^\circ$ field of view. Therefore $\mathbf d_p$ is only a stable image-plane direction coordinate. It is not a depth estimate and it is not a propagation-path label.

\subsection{Learning Objective}
The unlabeled pretraining set is
\begin{equation}
\cD_u=\{(I_i,\boldsymbol\rho_i)\}_{i=1}^{n_u},
\label{eq:unlabeled_set}
\end{equation}
and the labeled downstream set is
\begin{equation}
\cD_l=\{(I_i,\boldsymbol\rho_i,b_i^\star)\}_{i=1}^{n_l}.
\label{eq:labeled_set}
\end{equation}
Let $E_v$ and $E_w$ denote the vision and wireless encoders. Each encoder prepends a learned CLS token to its token sequence; after the final Transformer layer, the output at this CLS position is used as the global modality representation. We denote these CLS embeddings by $\mathbf z_i^v,\mathbf z_i^w\in\R^d$. CM-MAE pretraining has two objectives:
\begin{enumerate}
    \item \emph{Beam-profile soft alignment}: if two samples have similar normalized beam-power profiles, the model should assign them similar cross-modal retrieval probabilities.
    \item \emph{Masked conditional reconstruction}: hidden visual patches and hidden wireless cluster features should be reconstructed from the visible tokens of the available modality or modalities.
\end{enumerate}
For downstream beam-index classification, a task head $g_\phi$ is fitted on labeled source data:
\begin{equation}
(\theta^\star,\phi^\star)=
\arg\min_{\theta,\phi}
\mathbb E_{(I,\boldsymbol\rho,b^\star)\sim\cD_l}
\ell_{\rm cls}\!\left(g_{\phi}(E_{\theta}(I,\boldsymbol\rho)),b^\star\right).
\label{eq:downstream_problem}
\end{equation}
The fusion setting in this paper gives the classifier both the image and the contemporaneous 64-beam power vector at inference. This protocol evaluates cross-scenario multimodal representation transfer. It does not demonstrate proactive prediction before a sweep, nor does it show reduced beam-training overhead. The vision-only ablation is the corresponding camera-only setting.

\section{Evaluated CM-MAE Architecture}
\label{sec:method}

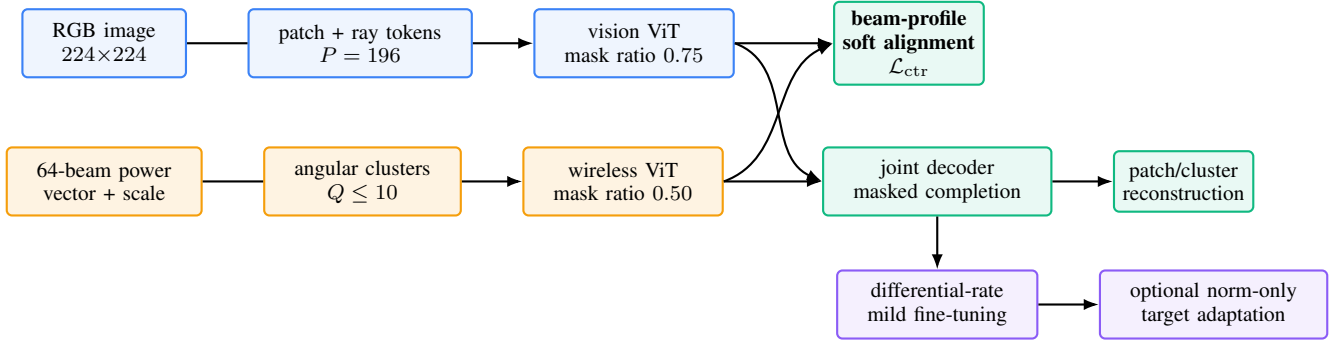
\begin{figure*}[t]
\centering
\begin{tikzpicture}[
    font=\footnotesize,
    node distance=6mm and 8mm,
    box/.style={rounded corners=2pt,draw,thick,align=center,minimum height=9mm,inner xsep=4mm},
    arr/.style={-{Latex[length=2mm]},thick},
    loss/.style={rounded corners=2pt,draw=joint,fill=joint!9,thick,align=center,minimum height=8mm}
]
\node[box,draw=vision,fill=vision!8] (img) {RGB image\\$224{\times}224$};
\node[box,draw=vision,fill=vision!8,right=of img] (vtok) {patch + ray tokens\\$P=196$};
\node[box,draw=vision,fill=vision!8,right=of vtok] (venc) {vision ViT\\mask ratio $0.75$};

\node[box,draw=wireless,fill=wireless!12,below=9mm of img] (pow) {64-beam power\\vector + scale};
\node[box,draw=wireless,fill=wireless!12,right=of pow] (ctok) {angular clusters\\$Q\leq10$};
\node[box,draw=wireless,fill=wireless!12,right=of ctok] (cenc) {wireless ViT\\mask ratio $0.50$};

\node[loss,draw=joint,fill=joint!9,right=13mm of venc] (ctr) {\textbf{beam-profile}\\\textbf{soft alignment}\\$\cL_{\rm ctr}$};
\node[box,draw=joint,fill=joint!9,right=13mm of cenc] (dec) {joint decoder\\masked completion};
\node[loss,right=of dec] (rec) {patch/cluster\\reconstruction};

\node[box,draw=adapt,fill=adapt!9,below=7mm of dec] (head) {differential-rate\\mild fine-tuning};
\node[box,draw=adapt,fill=adapt!9,right=of head] (tta) {optional norm-only\\target adaptation};

\draw[arr] (img)--(vtok)--(venc);
\draw[arr] (pow)--(ctok)--(cenc);
\draw[arr] (venc)--(ctr);
\draw[arr] (cenc.east) to[out=20,in=205] (ctr.west);
\draw[arr] (venc.east) to[out=-20,in=155] (dec.west);
\draw[arr] (cenc)--(dec);
\draw[arr] (dec)--(rec);
\draw[arr] (dec)--(head);
\draw[arr] (head)--(tta);
\end{tikzpicture}
\caption{Evaluated CM-MAE pipeline. Image patches and wireless angular clusters are encoded by separate Transformers, decoded jointly for masked reconstruction, and aligned globally by a beam-profile-guided soft contrastive loss.}
\label{fig:framework}
\end{figure*}

\subsection{Tokenization}
We use superscripts $v$ and $w$ for visual and wireless quantities. Patch indices are $p=1,\ldots,P$; wireless cluster indices are $q=1,\ldots,Q_i$ with $Q_i\leq Q_{\max}$.

\textbf{Vision tokens.}
A convolutional patch embedder $f_v$ maps image patch $I_{i,p}$ to width $d$. The token entering the vision encoder is
\begin{equation}
\mathbf x^{v}_{i,p}=f_v(I_{i,p})+W_r\mathbf d_p+\mathbf e_p,
\label{eq:vision_token}
\end{equation}
where $\mathbf e_p$ is a learned patch-position embedding and $W_r\mathbf d_p$ injects the approximate camera-ray coordinate from~\eqref{eq:ray}. This gives the model a consistent notion of image direction while avoiding unsupported depth or path claims.

\textbf{Wireless normalization and scale.}
The raw received-power vector is converted to dB and normalized per sample:
\begin{equation}
r_{i,b}=10\log_{10}(\rho_{i,b}+\epsilon),\qquad
p_{i,b}=\frac{r_{i,b}-m_i}{M_i-m_i+\epsilon},
\label{eq:db_norm}
\end{equation}
where $m_i=\min_b r_{i,b}$ and $M_i=\max_b r_{i,b}$. The normalized vector $\mathbf p_i=[p_{i,1},\ldots,p_{i,N_b}]^\top$ captures angular shape in $[0,1]^{N_b}$. Since this normalization removes absolute power, the scalar $M_i$ is also embedded as a wireless scale token
\begin{equation}
\mathbf x^w_{i,0}=W_s M_i.
\label{eq:scale_token}
\end{equation}

\textbf{Sparse angular clusters.}
Millimeter-wave power is often concentrated around a small number of angular lobes~\cite{heath}. The evaluated tokenizer therefore converts the dense 64-beam profile into a short cluster sequence. Local maxima of $\mathbf p_i$ above $\eta=0.15$ are detected, the strongest $Q_i\leq Q_{\max}=10$ peaks are retained, and the global maximum is used as a fallback if no local peak passes the threshold. For a retained peak at beam $b_{i,q}$, define energy $e_{i,q}=p_{i,b_{i,q}}$ and a local power-weighted angular centroid
\begin{equation}
\bar\theta_{i,q}=\frac{\sum_{b=b_{i,q}-\Delta_b}^{b_{i,q}+\Delta_b}p_{i,b}\theta_b}
{\sum_{b=b_{i,q}-\Delta_b}^{b_{i,q}+\Delta_b}p_{i,b}+\epsilon},
\quad \theta_b\in[-1,1],\;\Delta_b=2.
\label{eq:centroid}
\end{equation}
The local wireless feature $\mathbf a_{i,q}\in\R^6$ stores the normalized powers at $b_{i,q}-1,b_{i,q},b_{i,q}+1$ together with a zero imaginary channel, matching the real/imaginary feature shape used by the implementation. The cluster token is
\begin{equation}
\mathbf x^{w}_{i,q}=f_w(\mathbf a_{i,q})+
W_e[\log(e_{i,q}+\epsilon),\kappa_{i,q}]^\top+
\gamma(\bar\theta_{i,q}),
\label{eq:cluster_token}
\end{equation}
where $\kappa_{i,q}$ is a peak-quality indicator and $\gamma(\cdot)$ is a Fourier angular encoding~\cite{fourier}. Padded cluster slots are excluded by the attention mask. This tokenizer records angular location, local lobe shape, relative energy, and absolute scale, but it does not infer physical propagation paths.

\subsection{Dual Transformer Encoders and Joint Decoder}
The visual sequence is
\begin{equation}
\mathbf X_i^v=[\mathbf c^v,\mathbf x^v_{i,1},\ldots,\mathbf x^v_{i,P}],
\label{eq:vision_sequence}
\end{equation}
where $\mathbf c^v$ is the learned visual CLS token. The wireless sequence is
\begin{equation}
\mathbf X_i^w=[\mathbf c^w,\mathbf x^w_{i,0},\mathbf x^w_{i,1},\ldots,\mathbf x^w_{i,Q_i}],
\label{eq:wireless_sequence}
\end{equation}
where $\mathbf c^w$ is the wireless CLS token and $\mathbf x^w_{i,0}$ is the scale token from~\eqref{eq:scale_token}. Invalid padded cluster slots are masked out of attention.

The two streams use separate Vision Transformer encoders rather than shared weights. Each encoder is a pre-LayerNorm Transformer stack. For either modality, layer $\ell$ applies
\begin{align}
\mathbf H^{\ell+\frac12}&=\mathbf H^{\ell}+
\operatorname{MHSA}(\operatorname{LN}(\mathbf H^{\ell})),\nonumber\\
\mathbf H^{\ell+1}&=\mathbf H^{\ell+\frac12}+
\operatorname{FFN}(\operatorname{LN}(\mathbf H^{\ell+\frac12})).
\label{eq:transformer}
\end{align}
The final CLS outputs are
\begin{equation}
\mathbf z_i^v=\mathbf H^{v,L}_{i,0},\qquad
\mathbf z_i^w=\mathbf H^{w,L}_{i,0},
\label{eq:cls_embeddings}
\end{equation}
where position zero denotes the CLS token. The evaluated Base configuration uses width $d=768$, 12 encoder layers, 12 attention heads, and MLP ratio 4 in each encoder. This is a large model, roughly 232M parameters including the decoder and task head.

During pretraining, the joint decoder receives the concatenated encoded sequence
\begin{equation}
[\mathbf z^v_{\rm CLS},\mathbf z^v_{1:P},
\mathbf z^w_{\rm CLS},\mathbf z^w_{0},\mathbf z^w_{1:Q}]
\label{eq:decoder_seq}
\end{equation}
with modality-type embeddings. The decoder has eight Transformer layers, width 768, and 12 attention heads. Separate prediction heads map decoder outputs back to denormalized RGB patch vectors and six-dimensional local wireless features. The decoder is used only for self-supervised pretraining; downstream classification uses the unmasked encoders and a separate fusion head.

\subsection{Soft Contrastive Alignment Loss}
\label{sec:softloss}
\textbf{Why hard instance matching is insufficient.}
This is the soft contrastive alignment function referred to in the abstract. It is \emph{contrastive} because it compares visual and wireless CLS embeddings through a batch retrieval distribution, and it is \emph{soft} because its target is a probability distribution rather than a one-hot identity label. For a visual query $i$, hard instance matching would minimize $-\log\pi_{ii}^{v\rightarrow w}$ and treat every $j\neq i$ as a negative. This is inappropriate when two different samples have similar measured beam-power shapes. CM-MAE instead uses the wireless response itself to define how strongly sample $i$ should be associated with other samples in the batch.

\textbf{Wireless-derived soft targets.}
For batch size $B$, normalize each power profile as $\widetilde{\mathbf p}_i=\mathbf p_i/\|\mathbf p_i\|_2$ and define
\begin{equation}
d_{ij}=1-\widetilde{\mathbf p}_i^\top\widetilde{\mathbf p}_j,
\quad
q_{ij}=\frac{\mathbf{1}[j\neq i]\exp(-d_{ij}^2/(2\sigma_h^2))}
{\sum_{k\neq i}\exp(-d_{ik}^2/(2\sigma_h^2))},
\label{eq:soft_target}
\end{equation}
where $\sigma_h=0.5$. Since $\mathbf p_i$ is nonnegative and normalized, $d_{ij}$ measures angular-pattern discrepancy. The diagonal is set to zero in the evaluated implementation: $\cL_{\rm ctr}$ is used for non-self neighborhood learning, while the synchronized pair is still coupled through joint reconstruction. Smaller $\sigma_h$ concentrates probability on the nearest non-self profiles; larger $\sigma_h$ approaches a uniform distribution over $j\neq i$.

\textbf{Learned cross-modal distributions.}
Let $\mathbf u_i^v$ and $\mathbf u_i^w$ be $\ell_2$-normalized projections of the two CLS tokens and let $s_{ij}^{v\rightarrow w}=(\mathbf u_i^v)^\top\mathbf u_j^w/\tau$. The visual-to-wireless retrieval distribution is
\begin{equation}
\pi_{ij}^{v\rightarrow w}=
\frac{\exp(s_{ij}^{v\rightarrow w})}
{\sum_{k=1}^{B}\exp(s_{ik}^{v\rightarrow w})},
\label{eq:pred_dist}
\end{equation}
with $\tau=0.07$; $\boldsymbol\pi_i^{w\rightarrow v}$ is defined analogously. The symmetric soft alignment loss is
\begin{equation}
\cL_{\rm ctr}=-\frac{1}{2B}\sum_{i=1}^{B}\sum_{j=1}^{B}q_{ij}
\left(\log\pi_{ij}^{v\rightarrow w}+\log\pi_{ij}^{w\rightarrow v}\right).
\label{eq:contrastive}
\end{equation}

\textbf{What the function optimizes.}
For one direction, differentiation with respect to the similarity logit gives
\begin{equation}
\frac{\partial H(\mathbf q_i,\boldsymbol\pi_i^{v\rightarrow w})}
{\partial s_{ij}^{v\rightarrow w}}
=\pi_{ij}^{v\rightarrow w}-q_{ij}.
\label{eq:soft_gradient}
\end{equation}
Thus a nonpaired sample with large $q_{ij}$ is pulled closer until the learned probability matches the measured beam-profile affinity, while a distant sample with $q_{ij}\approx0$ remains a negative. The loss is therefore not a many-positive binary contrastive loss; it is distribution matching.

The distribution-matching interpretation follows from
\begin{align}
\cL_{\rm ctr}=C_q+\frac{1}{2B}\sum_{i=1}^{B}\big[
&\operatorname{KL}(\mathbf q_i\|\boldsymbol\pi_i^{v\rightarrow w})\nonumber\\
&+\operatorname{KL}(\mathbf q_i\|\boldsymbol\pi_i^{w\rightarrow v})\big],
\label{eq:kl_interpretation}
\end{align}
where $C_q=B^{-1}\sum_i H(\mathbf q_i)$ is independent of network parameters. Hence minimizing~\eqref{eq:contrastive} matches learned cross-modal retrieval neighborhoods to the wireless-response neighborhood. If all embeddings collapse, $\boldsymbol\pi_i$ becomes uniform; whenever $\mathbf q_i$ is nonuniform, the excess KL term is strictly positive. In this paper, ``physics-grounded'' means only that the target distribution is computed from measured propagation responses. It does not claim path recovery or universal electromagnetic invariance.

\subsection{Masking, Modality Dropout, and Reconstruction}

Masking follows the information density of the two modalities. Images are dense and spatially redundant, so vision uses a high mask ratio $r_v=0.75$ with contiguous row-major blocks of four patches. Wireless clusters are already sparse, so wireless uses a lower mask ratio $r_w=0.5$ over valid angular-cluster tokens. Let $m^v_{i,p},m^w_{i,q}\in\{0,1\}$ indicate whether a token's content is visible. Masked positions are not removed. Their content embedding is replaced by a learned mask token while their coordinate embedding remains:
\begin{equation}
\widetilde{\mathbf x}^{v}_{i,p}=m^v_{i,p}\mathbf x^v_{i,p}+
(1-m^v_{i,p})(\mathbf m_v+\mathbf e_p+W_r\mathbf d_p),
\label{eq:vision_mask}
\end{equation}
\begin{equation}
\widetilde{\mathbf x}^{w}_{i,q}=m^w_{i,q}\mathbf x^w_{i,q}+
(1-m^w_{i,q})(\mathbf m_w+\gamma(\bar\theta_{i,q})).
\label{eq:wireless_mask}
\end{equation}
Thus the model knows where a hidden image patch or angular cluster lies, but not its content. Keeping position information is important for the wireless stream: a position-free mask token would turn two missing clusters at different beam angles into the same decoder input.

Let $\Omega_i^v$ and $\Omega_i^w$ denote masked visual patches and masked valid wireless clusters. The two reconstruction losses are
\begin{align}
\cL_{\rm rec,v}&=\frac{1}{\sum_i|\Omega_i^v|}
\sum_i\sum_{p\in\Omega_i^v}\|\widehat{\mathbf t}_{i,p}^v-\mathbf t_{i,p}^v\|_2^2,\nonumber\\
\cL_{\rm rec,w}&=\frac{1}{\sum_i|\Omega_i^w|}
\sum_i\sum_{q\in\Omega_i^w}\|\widehat{\mathbf a}_{i,q}-\mathbf a_{i,q}\|_2^2.
\label{eq:reconstruction}
\end{align}
The joint decoder is used to make reconstruction conditional on both streams when both are present, so hidden visual patches or wireless clusters must be predicted from the visible context rather than from an isolated unimodal branch. This is a pretraining task design, not a claim that reconstruction loss alone guarantees transfer.

Modality dropout prevents the decoder from assuming that both modalities are always present. With probability $p_{\rm md}=0.1$ the vision branch is dropped, and with another $0.1$ the wireless branch is dropped. Let $a_v,a_w\in\{0,1\}$ denote branch presence. The implemented pretraining loss is
\begin{align}
\cL_{\rm pre}={}&a_v\cL_{\rm rec,v}+a_w\beta_w\cL_{\rm rec,w}
+a_va_w\lambda_{\rm ctr}\cL_{\rm ctr},\nonumber\\
&\beta_w=1,\qquad \lambda_{\rm ctr}=0.2.
\label{eq:gated_total}
\end{align}
The product gate removes the alignment term whenever either CLS embedding is absent; reconstruction remains active for whichever branch is present.

\section{Learning and Adaptation Procedure}
\label{sec:training}

\subsection{Stage 1: Self-Supervised Pretraining}
Pretraining uses paired but unlabeled samples $(I_i,\boldsymbol\rho_i)$ from the source scenarios. For each mini-batch, the tokenizer builds visual patch tokens, the scale token, and wireless angular-cluster tokens; independent masks and modality-dropout indicators are sampled; the two encoders and joint decoder are optimized with $\cL_{\rm pre}$ in~\eqref{eq:gated_total}. The beam-index label $b_i^\star$ is not used in this stage. The soft contrastive loss is active only when both modality branches are present, and reconstruction is active for every available branch. Checkpoint selection is performed on the source validation scenario after downstream adaptation, not by target-scenario labels.

\subsection{Stage 2: Source-Supervised Adaptation}
After pretraining, no reconstruction mask is applied. The unmasked encoders produce $\mathbf z^v_i$ and $\mathbf z^w_i$. In fusion mode, a projection of the full normalized power vector $g_p(\mathbf p_i)$ is added to the wireless CLS feature. A two-layer bidirectional residual attention module exchanges information between the visual and wireless global features, and an MLP predicts one of the $N_b=64$ beam indices.

Uniform learning rates caused large representation drift in our cross-scenario runs. The final adaptation rule therefore uses two optimizer groups. If $\theta_e$ denotes pretrained encoder parameters and $\theta_h$ denotes newly initialized fusion and classifier parameters, the updates are
\begin{equation}
\theta_e^{t+1}=\theta_e^t-\eta_e\nabla_{\theta_e}\cL_{\rm cls},\quad
\theta_h^{t+1}=\theta_h^t-\eta_h\nabla_{\theta_h}\cL_{\rm cls},
\label{eq:mild_update}
\end{equation}
with $\eta_e=3\times10^{-6}$ and $\eta_h=10^{-3}$. AdamW uses cosine decay, five-epoch warmup, weight decay 0.1, label smoothing 0.05, head dropout 0.15, gradient clipping at 1, and 80 epochs with effective batch size 512.

The peak head rate is about $333$ times the encoder rate. The intended two-timescale behavior is simple: adapt the task-specific head quickly while limiting movement away from the pretrained encoder.

\subsection{Stage 3: Optional Test-Time Adaptation}
For optional transductive adaptation, only affine parameters of normalization layers are updated on unlabeled target-scenario batches by minimizing
\begin{equation}
\cL_{\rm TTA}=-\frac{1}{B}\sum_{i=1}^{B}\sum_{b=1}^{N_b}
\widehat p_{i,b}\log\widehat p_{i,b}.
\label{eq:tta}
\end{equation}
Adam uses learning rate $5\times10^{-4}$ for 30 steps. Adaptation is performed separately for each target scenario and the source weights are restored before moving to another scenario. This stage is transductive: it assumes access to unlabeled target-scenario batches, and labels are used only after adaptation to report accuracy. We therefore report the source-only mild-FT result and the mild-FT+TTA result separately.

\subsection{Out-of-Scope Privileged Geometry}
We reserve the name CM-MAE-W for a possible geometry-privileged extension with complex wideband CFR, 3-D angle--delay clusters, geometry-guided masks, and a supervised correspondence head. That version is not part of the evaluated method because the reported DeepSense experiments do not use ray-traced path labels, calibrated depth, or patch--path association. A future dataset with such labels could add entropic optimal-transport correspondence with dustbins~\cite{cuturi}, but that would be a different supervision regime. The results below therefore evaluate only the real-data RGB-plus-measured-power setting defined in Sections~\ref{sec:setup}--\ref{sec:training}.

\section{Experiments}
\label{sec:experiments}

\subsection{Setup}

\noindent\textbf{Dataset and protocol.} DeepSense~6G Scenarios~1--8 contain real-world vehicle-to-infrastructure (V2I) measurements at 60~GHz with synchronized RGB images ($224{\times}224$) and 64-beam power vectors. Each scenario corresponds to a different geographic location. We split each scenario 70/15/15 by \texttt{seq\_index} to avoid leakage from consecutive frames of the same vehicle. Pretraining uses unlabeled pairs from Sc1--5; fine-tuning uses labeled data from Sc1--4; Sc5 serves as validation; Sc6--8 are held out as the unseen test set. All non-TTA accuracies use the \emph{test split} of each scenario without target labels during training or model selection. The TTA row additionally uses unlabeled target-scenario batches for entropy minimization and is marked separately as transductive. The test sets contain 488 samples (Sc5), 254 (Sc6), 159 (Sc7), and 652 (Sc8).

\noindent\textbf{Implementation.} We use ViT-Base encoders (two streams, $\sim$232~M parameters total). The vision stream is initialized from ImageNet-pretrained ViT-B/16; the CSI stream is trained from scratch. Pretraining uses AdamW (peak LR $1.2{\times}10^{-4}$, weight decay $0.05$, batch size 192, cosine schedule with 10-epoch warmup) on a single A100 GPU with automatic mixed precision. \emph{All} models in this paper---the main model and every ablation variant---are pretrained for 160 epochs under the identical protocol, so that differences reflect the design choices rather than training duration. The best checkpoint is selected by validation linear probe. Fine-tuning runs on one A100. Images use ImageNet mean/std normalization; beam angles are mapped to $[-1,1]$.

\noindent\textbf{Baselines.} We compare with the official ASU ResNet-18 vision-only baseline~\cite{visionposition} evaluated under the same cross-scenario protocol. To position CM-MAE among mainstream SSL paradigms, we further pretrain three representative baselines under the identical protocol: CLIP-style hard InfoNCE alignment without masking (CLS-InfoNCE), soft-contrastive alignment without masking (CLS-SoftCtr), and masked reconstruction without alignment (MAE). We also report several CM-MAE variants that isolate the contribution of each component: linear probe (frozen encoder), full fine-tuning (uniform LR $10^{-5}$), single-modality mild fine-tuning (vision-only and CSI-only), and mild fine-tuning with and without TTA.

\subsection{Main Results}

Table~\ref{tab:main} reports cross-scenario Top-1 accuracy. CM-MAE mild fine-tuning reaches \textbf{77.4\%} on Sc6--8, outperforming the ASU baseline by $+24.9$~pp. TTA adds another $+1.3$~pp to reach 78.7\%. The gain is largest on the hardest scenarios: Sc6 (low-light residential, $+22.5$~pp) and Sc7 (dense urban blockage, $+31.6$~pp).

\begin{table}[t]
\centering
\caption{Cross-scenario Top-1 accuracy (\%) on DeepSense~6G. Labels on Sc1--4; Sc5 validation; Sc6--8 unseen test. CM-MAE variants use fusion unless noted.}
\label{tab:main}
\setlength{\tabcolsep}{3.5pt}
\small
\begin{tabular}{lccccc}
\toprule
\textbf{Method} & \textbf{Sc5} & \textbf{Sc6} & \textbf{Sc7} & \textbf{Sc8} & \textbf{Avg6--8} \\
\midrule
ASU ResNet-18~\cite{visionposition} & 58.27 & 41.67 & 43.90 & 71.92 & 52.50 \\
\midrule
Linear probe & 65.57 & 21.65 & 30.82 & 62.58 & 38.35 \\
Full FT & 85.66 & 28.74 & 38.99 & 86.35 & 51.36 \\
Vision-only (mild) & 43.03 & 26.77 & 21.38 & 40.64 & 29.60 \\
CSI-only (mild) & 81.35 & 18.90 & 22.64 & 63.65 & 35.06 \\
\midrule
\textbf{Mild-FT (ours)} & \textbf{94.26} & \textbf{64.17} & \textbf{75.47} & \textbf{92.48} & \textbf{77.38} \\
\textbf{Mild-FT + TTA} & 93.44 & 66.54 & 76.73 & 92.79 & \textbf{78.69} \\
\bottomrule
\end{tabular}
\end{table}

\noindent Key observations:

\noindent(i) \textbf{Fusion is essential under the diagnostic protocol.} Neither single modality generalizes across scenarios: vision-only mild-FT reaches only 29.6\% on Sc6--8 (below the ASU vision baseline), and CSI-only reaches 35.1\%. Their fusion reaches 77.4\%, showing that the learned visual representation and contemporaneous power-vector representation are complementary in this representation-transfer setting. Because the fusion classifier receives the full 64-beam power vector at inference, this row should not be read as a camera-only proactive beam-prediction result.

\noindent(ii) \textbf{Full fine-tuning destroys generalization.} Despite a high Sc5 accuracy of 85.7\%, full FT collapses to 51.4\% on Sc6--8---at the level of the ASU baseline---because a uniform learning rate overwrites the cross-modal features with scenario-specific cues from Sc1--4. The high Sc5 accuracy is misleading because Sc5 shares the visual domain of the training scenarios.

\noindent(iii) \textbf{TTA helps where it matters, when target batches are available.} Adaptation improves the hardest scenarios Sc6/Sc7 by $+2.4$/$+1.3$~pp while leaving the already-easy Sc8 nearly unchanged, consistent with the entropy-minimization objective targeting uncertain predictions. This gain is not source-only; it uses unlabeled target data and is therefore reported as an optional transductive result.

\subsection{Ablation Studies}
\label{sec:ablation}

We isolate the three most decisive design choices: the pretraining objective, the fine-tuning strategy, and the masking design.

\noindent\textbf{(A) Pretraining objectives.}
We compare six pretraining variants under an identical protocol: ImageNet initialization, Sc1--5 unlabeled data, 160 epochs, batch size 192, evaluated by the \emph{same} linear-probe procedure (a linear classifier trained on Sc1--4 features and tested on the held-out test split of each scenario; final epoch-159 checkpoint). The variants factorize along two axes---\emph{masking} (none vs.\ block masking with reconstruction) and \emph{alignment} (none, hard InfoNCE~\cite{clip}, or our soft beam-correlation targets)---and include the mainstream SSL paradigms as baselines.

Table~\ref{tab:abl_obj} yields several findings. First, reconstruction alone (MAE rows) underperforms any alignment-based objective, and block masking without alignment collapses entirely---the contrastive term is the primary representation-learning signal. Second, \emph{soft targets beat hard targets at every masking setting}: CLS-SoftCtr $>$ CLS-InfoNCE ($+2.6$~pp) and CM-MAE-U $>$ MAE+InfoNCE ($+5.9$~pp), confirming that physics-grounded partial credit matters because many samples share similar propagation patterns.

Third, and most instructive, the ranking \emph{reverses} between frozen evaluation and fine-tuning. Under a linear probe, CLS-SoftCtr (38.1\%) leads our full model (31.7\%) because global CLS alignment produces readily separable features. But once the encoder is unfrozen during mild fine-tuning, masked pretraining pulls decisively ahead (Table~\ref{tab:abl_ft_obj}): CM-MAE-U reaches \textbf{45.0\%} on Sc6--8 versus \textbf{37.3\%} for CLS-SoftCtr, a $+7.7$~pp gain concentrated in the hardest unseen scenario (Sc8: 80.1\% vs 61.3\%). Masked completion forces the encoder to model \emph{within-image spatial structure}---which patches belong to the same vehicle, which are occluded---and these local features are exactly what mild fine-tuning exploits when adapting to a new scenario. Pure CLS alignment, lacking this local grounding, overfits the global statistics of the training scenarios and transfers poorly. Masked completion is therefore best understood as a regularizer whose benefit materializes at adaptation time, not at the frozen-feature level.

\begin{table}[t]
\centering
\caption{Frozen linear probe vs.\ mild fine-tuning (Avg6--8 Top-1, fusion) for the two strongest pretraining variants. Masked pretraining trails when frozen but leads after adaptation.}
\label{tab:abl_ft_obj}
\setlength{\tabcolsep}{4pt}
\small
\begin{tabular}{lcc}
\toprule
\textbf{Pretraining} & \textbf{Linear probe} & \textbf{Mild FT} \\
\midrule
CLS-SoftCtr (no mask) & \textbf{38.14} & 37.34 \\
\textbf{CM-MAE-U (mask + soft)} & 31.71 & \textbf{45.04} \\
\bottomrule
\end{tabular}
\end{table}

\begin{table}[t]
\centering
\caption{Ablation on pretraining objectives. All variants pretrained 160 epochs (ImageNet init, Sc1--5, batch 192), linear probe on the test split (fusion features).}
\label{tab:abl_obj}
\setlength{\tabcolsep}{2.5pt}
\footnotesize
\begin{tabular}{llccccc}
\toprule
\textbf{Variant} & \textbf{Mask.} & \textbf{Align.} & \textbf{Sc6} & \textbf{Sc7} & \textbf{Sc8} & \textbf{Avg} \\
\midrule
CLS-InfoNCE & none & hard & 21.26 & 32.08 & 53.37 & 35.57 \\
CLS-SoftCtr & none & soft & 27.56 & 27.67 & 59.20 & \textbf{38.14} \\
MAE & block & -- & 11.02 & 11.32 & 52.30 & 24.88 \\
MAE & rand. & -- & 12.20 & 16.98 & 60.43 & 29.87 \\
MAE\,+\,InfoNCE & block & hard & 11.02 & 14.47 & 51.84 & 25.78 \\
\textbf{CM-MAE-U} & block & soft & 14.17 & 17.61 & 63.34 & 31.71 \\
\bottomrule
\end{tabular}
\end{table}

\noindent\textbf{(B) Fine-tuning strategy.}
Starting from the \emph{same} pretrained encoder, we compare three adaptation strategies, all evaluated test-only on Sc6--8 (Table~\ref{tab:abl_ft}): linear probing (frozen encoder, head only), full fine-tuning (uniform LR $10^{-5}$ across all parameters), and mild fine-tuning (differential LR). The trend is striking: linear probing is stable but weak (38.4\%); full fine-tuning fits the source scenarios well but forgets pretrained knowledge and fails to transfer (51.4\%); mild fine-tuning preserves the cross-modal features and reaches 77.4\%, a $+26.0$~pp jump over full FT. This validates our central claim that \emph{how} the pretrained model is adapted matters as much as the pretraining itself.

\begin{table}[t]
\centering
\caption{Ablation on the fine-tuning strategy (same pretrained encoder, fusion).}
\label{tab:abl_ft}
\setlength{\tabcolsep}{5pt}
\small
\begin{tabular}{lccccc}
\toprule
\textbf{Strategy} & \textbf{Sc5} & \textbf{Sc6} & \textbf{Sc7} & \textbf{Sc8} & \textbf{Avg6--8} \\
\midrule
Linear probe (frozen) & 65.57 & 21.65 & 30.82 & 62.58 & 38.35 \\
Full FT (uniform LR) & 85.66 & 28.74 & 38.99 & 86.35 & 51.36 \\
\midrule
\textbf{Mild FT (ours)} & \textbf{94.26} & \textbf{64.17} & \textbf{75.47} & \textbf{92.48} & \textbf{77.38} \\
\bottomrule
\end{tabular}
\end{table}

\noindent\textbf{(C) Masking design.}
All masking variants share the full objective ($\lambda_{\text{ctr}}{=}0.2$) and the protocol of (A), changing one design factor at a time from the base configuration ($r_v{=}0.75$ block, $r_c{=}0.5$, modality dropout $p{=}0.1$). Table~\ref{tab:abl_mask} varies the vision mask ratio $r_v$, the CSI cluster mask ratio $r_c$, the vision masking type (contiguous $4{\times}1$ blocks vs.\ uniform random), and modality dropout.

Vision masking follows an inverted-U pattern: the base ratio $r_v{=}0.75$ (31.7\%) outperforms both lighter ($r_v{=}0.5$: 28.8\%) and heavier ($r_v{=}0.9$: 27.4\%) masking, because too-few visible tokens prevent the model from inferring scene geometry while too-many make the reconstruction task trivially easy. CSI masking shows a similar pattern, with $r_c{=}0.75$ hurting most ($-6.3$~pp) since aggressive masking destroys the few salient beam clusters. Block masking beats uniform random masking by $+2.5$~pp: contiguous blocks force the model to fill in larger semantic regions (e.g., an occluded vehicle) from cross-modal context, whereas random tokens can be solved by local interpolation. The no-dropout row is higher in this frozen linear-probe table (34.02\% vs.\ 31.71\%), so Table~\ref{tab:abl_mask} does not by itself prove a transfer-accuracy gain from modality dropout. We keep dropout in the reported model as a conservative robustness-oriented choice for missing or unreliable modalities, and leave explicit corruption tests as future validation.

\begin{table}[t]
\centering
\caption{Ablation on the masking design (full objective, 160 epochs, batch 192), linear probe on the test split (fusion features).}
\label{tab:abl_mask}
\setlength{\tabcolsep}{3pt}
\footnotesize
\begin{tabular}{lccccc}
\toprule
\textbf{Variant} & \textbf{Sc6} & \textbf{Sc7} & \textbf{Sc8} & \textbf{Avg6--8} \\
\midrule
Base ($r_v{=}.75$, $r_c{=}.5$, $p_{md}{=}.1$) & 14.17 & 17.61 & 63.34 & 31.71 \\
$r_v{=}0.50$ & 10.24 & 15.72 & 60.43 & 28.80 \\
$r_v{=}0.90$ & 15.35 & 19.50 & 47.24 & 27.36 \\
$r_c{=}0.25$ & 12.60 & 15.72 & 67.64 & 31.99 \\
$r_c{=}0.75$ & 8.27 & 6.29 & 61.81 & 25.46 \\
Random masking (no blocks) & 12.20 & 14.47 & 60.89 & 29.19 \\
No modality dropout ($p_{md}{=}0$) & 17.32 & 16.35 & 68.40 & \textbf{34.02} \\
\bottomrule
\end{tabular}
\end{table}

\noindent\textbf{What did not work.}
A domain-adversarial scenario-classification head with a gradient-reversal layer~\cite{dann} was added to the fused CLS features to encourage scenario-invariant representations. However, the adversarial loss spiked repeatedly during pretraining, causing gradient explosion and NaN; disabling it ($\lambda_{\text{da}}{=}0$) consistently gave the best result. We also found that continuing SSL pretraining on all eight scenarios with a freshly reset optimizer degraded Sc6--8 accuracy (to $\sim$49\%), presumably because resetting Adam momentum and the learning-rate schedule disrupted the already-converged features.

\subsection{Comparison with Published Methods}

Table~\ref{tab:sota} places CM-MAE among recent published methods. The table is a positioning summary, not a protocol-matched leaderboard: most prior numbers are in-distribution (same-scenario train/test), whereas our main protocol requires generalization to unseen geographic locations. The only directly matched comparison in this table is the ASU ResNet cross-scenario reference.

\begin{table}[t]
\centering
\caption{Positioning against published methods. Cross-scenario: labels on Sc1--4, test on Sc6--8; in-distribution: same-scenario train/test.}
\label{tab:sota}
\setlength{\tabcolsep}{4pt}
\small
\begin{tabular}{llcc}
\toprule
\textbf{Method} & \textbf{Setting} & \textbf{Metric} & \textbf{Result} \\
\midrule
ASU ResNet~\cite{visionposition} & Cross-scenario & Top1 Avg6--8 & 52.5\% \\
BeamLLM~\cite{zheng2025beamllm} & In-distribution & Top1 & 61.0\% \\
Semantic BF~\cite{raha2024advancing} & Scenario-specific & Top1 & -- \\
Occlusion-Aware~\cite{orimogunje2026occlusion} & In-distribution & Top1 & 50.9\% \\
\midrule
\textbf{CM-MAE Mild-FT} & \textbf{Cross-scenario} & \textbf{Top1 Avg6--8} & \textbf{77.4\%} \\
\textbf{CM-MAE + TTA} & \textbf{Cross-scenario} & \textbf{Top1 Avg6--8} & \textbf{78.7\%} \\
\bottomrule
\end{tabular}
\end{table}

\section{Analysis and Discussion}
\label{sec:analysis}

\subsection{Per-Scenario Analysis}

Sc8 (highway/urban canyon; 652 test samples) is the easiest scenario at 92.5\%, reflecting its larger dataset and more structured propagation environment. Sc6 (254 test samples, low-light residential) and Sc7 (159 samples, dense urban with blockage) are harder but still improve by $+22.5$ and $+31.6$~pp respectively over the ASU baseline. The Sc5 validation accuracy of 94.3\% is high because Sc5 shares the visual domain of Sc1--4 (similar lighting and architecture); we caution that Sc5 accuracy is not a reliable proxy for cross-scenario generalization, and the true measure is Sc6--8.

The single-modality results reveal an interesting asymmetry: CSI-only mild-FT achieves high Sc5 accuracy (81.4\%) but poor Sc6/7 transfer (18.9\% / 22.6\%), while vision-only is weak everywhere (Sc5 only 43.0\%). This suggests that the raw power vector carries strong scenario-specific information that does not transfer, while the pretrained vision stream needs the CSI alignment target to learn transferable features---their combination is what enables generalization.

\subsection{Effect of the Soft Contrastive Targets}

The choice of beam-power correlation as the source of soft targets is important. Unlike self-supervised contrastive learning that uses image augmentations to define positives (irrelevant for CSI), our targets are grounded in the physical signal: two samples with similar beam-power profiles genuinely share propagation characteristics. The Gaussian kernel with $\sigma_h{=}0.5$ provides a smooth similarity landscape that tolerates the noise in real power measurements. Our ablation (Table~\ref{tab:abl_obj}) confirms that this signal is necessary: reconstruction alone leads to vision-stream collapse, while adding the contrastive term yields a representation that continues improving throughout pretraining.

\subsection{Limitations and Future Work}

\begin{itemize}[leftmargin=*]
    \item \textbf{Path-resolved correspondence.} Our current alignment operates at the sample level (CLS tokens). A finer objective---optimal-transport matching between image patches and beam clusters with asymmetric geometry-guided masking (the CM-MAE-W variant in our formulation)---requires ray-traced path labels (AoA/AoD, path powers) that are only available from simulators such as Wireless InSite. Evaluating it on synthetic data and transferring the representation to real measurements is ongoing work.
    \item \textbf{Temporal modeling.} Predictions are single-frame; incorporating short image--CSI sequences (as in the DeepSense challenge task) could add robustness against transient occlusions.
    \item \textbf{Pretraining scale and evaluation.} Our negative result on all-scenario continued pretraining suggests that optimizer-state handling and learning-rate scheduling for continued SSL deserve further study. Evaluation on the hidden DeepSense challenge test set would also provide an unbiased external comparison.
    \item \textbf{Computational cost.} The dual-stream ViT-Base architecture is large ($\sim$232~M parameters); distillation into a smaller model would aid deployment.
\end{itemize}

\section{Conclusion}
\label{sec:conclusion}

We presented CM-MAE, a physics-guided cross-modal masked autoencoder for vision--wireless representation transfer in 6G beam classification. Self-supervised alignment via beam-power soft contrastive learning improves matched transfer ablations, and a mild differential-LR fine-tuning strategy preserves pretrained cross-modal features better than uniform full fine-tuning. Under our DeepSense~6G cross-scenario protocol, CM-MAE reaches 77.4\% Top-1 on Sc6--8, a $+24.9$~pp gain over the protocol-matched ASU ResNet reference; optional transductive normalization adaptation reaches 78.7\%. Because the fusion setting uses the contemporaneous 64-beam power vector, these results support cross-scenario representation transfer rather than proactive reduced-sweep beam prediction.

\end{document}